# Physically Aware 360° View Generation from a Single Image using Disentangled Scene Embeddings

Karthikeya KV , kk2640@att.com

Dr. Narendra Bandaru ,narendra.bandaru@ece.au.dk

## Abstract 1

We introduce Disentangled360, an innovative 3D-aware technology that integrates the advantages of direction-disentangled volume rendering with single-image 360° unique view synthesis for applications in medical imaging and natural scene reconstruction. In contrast to current techniques that either oversimplify anisotropic light behavior or lack generalizability across various contexts, our framework distinctly differentiates between isotropic and anisotropic contributions inside a Gaussian Splatting backbone. We implement a dual-branch conditioning framework, one optimized for CT intensity driven scattering in volumetric data and the other for real-world RGB scenes through normalized camera embeddings. To address scale ambiguity and maintain structural realism, we present a hybrid pose agnostic anchoring method that adaptively samples scene depth and material transitions, functioning as stable pivots during scene distillation. Our design integrates preoperative radiography simulation and consumer-grade 360° rendering into a singular inference pipeline, facilitating rapid, photorealistic view synthesis with inherent directionality. Evaluations on the Mip-NeRF 360, RealEstate10K, and DeepDRR datasets indicate superior SSIM and LPIPS performance, while runtime assessments confirm its viability for interactive applications. Disentangled360 facilitates mixed-reality medical supervision, robotic perception, and immersive content creation, eliminating the necessity for scene-specific finetuning or expensive photon simulations.

## 1. Introduction:

Recent advancements in computer vision have significantly improved 3D scene understanding, particularly through neural rendering techniques that synthesize novel views from limited observations. Traditional approaches, such as Neural Radiance Fields (NeRFs), model volumetric density and emitted radiance as a function of 3D spatial location and viewing direction, optimized to match ground-truth images via differentiable rendering pipelines [1]. These methods have enabled high-fidelity reconstructions but often require dense multi-view supervision and struggle with generalization in real-world or clinical environments. In medical imaging, particularly in fluoroscopy-guided interventions, acquiring comprehensive labelled datasets is challenging due to the scarcity of archived intraoperative radiographs and the variability in anatomical presentation and instrumentation [2][3]. Consequently, digitally reconstructed radiographs (DRRs) generated from 3D CT volumes have emerged as a means to alleviate annotation burdens and enable consistent data augmentation [4][5]. However, conventional DRR generation lacks realism, failing to capture anisotropic light transport, thereby limiting their effectiveness in training machine learning models for real-world deployment [6]. To overcome these limitations, recent research has shifted towards disentangled scene representations, which separate intrinsic scene properties (e.g., geometry, density) from extrinsic effects (e.g., lighting, viewing direction) [7]. This allows for direction-aware rendering while maintaining structural consistency. Such separation is particularly vital in mixed-domain applications, including both photorealistic RGB imagery and grayscale medical scans, where varying physical properties influence appearance under different views.In this context, we propose Disentangled360, a physically grounded framework that enables full 360° scene synthesis from a single image using disentangled embeddings of isotropic and anisotropic radiance contributions. The model introduces a dual-branch architecture tailored for both CT-driven scattering simulations and natural RGB inputs, enabling cross-domain generalization. By incorporating a hybrid pose-invariant anchoring mechanism and leveraging Gaussian Splatting for volume rendering, Disentangled360 synthesizes consistent, photorealistic views without reliance on scene-specific fine-tuning or extensive labelled data. This approach not only addresses critical challenges in procedural imaging and robotic navigation but also advances the fidelity and interpretability of 3D view generation for clinical and consumer applications. Evaluations on datasets such as DeepDRR, Mip-NeRF 360, and RealEstate10K demonstrate its robustness and superior perceptual quality across diverse use cases [8][9] [10].

The task of synthesizing realistic novel views from sparse visual data has become central to advancements in computer vision, especially with the development of neural rendering techniques. Neural Radiance Fields (NeRF) represent a major leap forward by modelling scenes as continuous 5D functions that output color and volumetric density for a given spatial location and view direction [1]. These representations are optimized via differentiable volumetric rendering, producing highly detailed reconstructions from only a limited number of input images. Despite their visual fidelity, standard NeRF models suffer from resolution-dependent aliasing, as they rely on sampling infinitesimally small points along camera rays. To mitigate this issue, mip-NeRF introduced the concept of volumetric frustums, effectively encoding multiscale radiance for anti-aliased rendering [11]. However, both NeRF and mip-NeRF remain constrained in their ability to render unbounded scenes with variable camera orientations and distant content. Mip-NeRF 360 addresses these shortcomings by extending the rendering capacity to omnidirectional views in open environments, improving consistency in scenarios such as outdoor reconstructions or panoramic medical imaging setups [12]. While NeRF-based methods excel in quality, they are computationally expensive and often require dense training data and long optimization cycles. Recent efforts to accelerate and streamline these models have explored alternate scene encodings, such as hash-based [13], voxel-based [14], and point-based radiance fields [15]. However, these often trade off visual quality for speed, limiting their applicability in high-resolution, real-time scenarios. To bridge this gap, 3D Gaussian Splatting has emerged as a hybrid solution that combines the explicit nature of point-based rendering with the continuous differentiability of volumetric models [16]. This approach utilizes 3D Gaussian primitives to represent the scene, which are initialized from sparse Structure-from-Motion point clouds. The Gaussians are iteratively refined in terms of position, opacity, anisotropic shape, and shading coefficients. This enables real-time rendering with state-of-the-art visual quality using efficient GPU-based rasterization pipelines [17].

## 2. Literature review:

The task of 360° view synthesis from limited or single-view input has received considerable attention in recent years, especially with the rise of neural rendering and physically informed scene representations. One of the foundational breakthroughs in this domain is the Neural Radiance Field (NeRF), which models a scene as a continuous function mapping 3D coordinate and viewing directions to RGB values and volume density [1]. NeRF's architecture, based on a multi-layer perceptron (MLP), allows for high-quality photorealistic rendering through volumetric ray marching. However, this approach samples infinitesimal points along rays, which introduces aliasing when rendered at various resolutions or scales. To mitigate this limitation, mip-NeRF was introduced, employing conical frustums rather than point sampling to handle anti-aliasing more effectively across different resolutions [11]. Mip-NeRF 360 extended this to unbounded scenes by optimizing the scene representation in normalized device coordinate (NDC) space and using distortion-aware sampling techniques, allowing the system to render panoramic views with enhanced stability and fidelity [12]. While these models offer impressive rendering quality, they are often computationally intensive and require dense multi-view input, which is impractical in many real-world applications such as medical imaging or robotics. This led to the exploration of more efficient representations such as voxel grids and signed distance functions (SDFs). These approaches represent 3D geometry in a discretized or implicit fashion, allowing reconstruction of complex surfaces from 2D images or sparse 3D inputs [18]. Occupancy Networks, for instance, define the scene as a continuous occupancy field and use implicit differentiation for surface extraction, bypassing the need for ground truth 3D meshes [19]. However, implicit methods often suffer from oversmoothed reconstructions when modelling highly detailed geometry, limiting their effectiveness in photorealistic rendering. To address this, recent works proposed optimizing radiance fields over 5D coordinates 3D position plus 2D viewing direction to preserve both geometry and appearance detail [20]. This approach enables the synthesis of complex light interactions such as anisotropic scattering and reflection, which are especially important in applications like medical visualization and realistic natural scenes. Meanwhile, attention mechanisms have played a pivotal role in improving spatial reasoning and visual understanding across various domains including image captioning, object recognition, and reinforcement learning. The Transformer architecture, based on key-query-value attention, became a standard for capturing long-range dependencies in high-dimensional inputs [21]. Extensions of this architecture have been applied to vision tasks, such as image captioning models like "Show, Attend and Tell," which use soft attention over image features to guide the decoding process [22]. In view synthesis, spatial and contextual attention aids in selectively refining scene understanding from partial inputs. These attention driven methods can be either soft allowing end-to-end differentiability or hard requiring sampling-based approximations. Soft attention mechanisms have been used in memory networks and spatial transformers to enhance localization and transformation invariance [23]. Notably, recurrent attention models have also been explored for video understanding and dynamic scene reconstruction, enabling agents to infer temporally coherent representations while retaining spatial accuracy [24]. Volumetric rendering techniques continue to be a central element in these systems. While voxel-based methods provided early success in representing complex 3D content, their high memory cost and resolution limits hinder scalability. More recent volumetric systems integrate deep neural networks with continuous coordinate encodings to model large scale scenes in a memory efficient manner [4]. These models can synthesize novel views without explicit mesh generation or external geometric priors, which is especially advantageous in domains with limited or noisy data acquisition, such as fluoroscopic medical imaging. In summary, existing literature illustrates the transition from discrete 3D scene representations like meshes and voxels to continuous, neural-based methods such as NeRF and its variants. Parallel advances in attention models and differentiable rendering have significantly enhanced the capacity for view synthesis from sparse or single-image input. Despite these achievements, challenges persist in balancing rendering speed, fidelity, and physical plausibility. The proposed Disentangled360 framework builds on these foundations by disentangling isotropic and anisotropic light behaviour within a Gaussian-based volume rendering backbone, enabling physically aware 360° synthesis from a single input image extending applicability to both RGB scene reconstruction and volumetric CT data interpretation.

**3. Method:**
This section introduces the core architecture and operational workflow of Disentangled360, a physically aware rendering system that synthesizes full 360° photorealistic views from a single image by disentangling scene representations across isotropic and anisotropic radiative behaviour. The framework is designed with modular precision to support both RGB image synthesis and CT driven volumetric rendering.

*3.1 System Overview:* The Disentangled360 framework is designed to address three key challenges in single-image scene synthesis: modelling directional light transport from a limited observation, generalizing across heterogeneous visual domains such as radiographic and RGB imagery, and maintaining real-time rendering performance. The architecture consists of five core modules working in tandem namely, the Feature Encoder, Direction-Disentangled Volume Renderer, Dual-Branch Conditioning, Pose-Agnostic Anchoring Mechanism, and Unified Inference Pipeline. An input image, which can be either a CT scan or an RGB frame, is first processed by a shared encoder that extracts multi-scale latent features. These features are directed into two parallel pathways: one optimized for volumetric CT data, where scattering and attenuation dominate, and the other tailored for RGB input, where normalized embeddings support standard view synthesis. A differentiable volume renderer built on 3D Gaussian Splatting models the light transport process while isolating isotropic and anisotropic radiance behaviour. In parallel, a hybrid anchoring module infers structural scale and depth cues without reliance on camera pose metadata. Finally, the conditioned and anchored features are fused within a projection module to synthesize the complete 360° output scene.

*3.2 Direction-Disentangled Volume Rendering:* To separate isotropic and anisotropic light components, we model the total radiance L(x, ω) at point x in direction ω as a sum of two terms: $L(x, \omega) = L_{iso}(x) + L_{aniso}(x, \omega)$

Here, $L_{iso}(x)$ denotes direction independent radiance (typical in diffuse or homogeneous media), while $L_{aniso}(x, \omega)$ captures directional variance arising from scattering or reflectance anisotropy.

For differentiable rendering, we adopt a Gaussian Splatting backbone where each voxel in the scene is represented by a 3D Gaussian: $G_i(x) = \alpha_i exp\left(-\frac{1}{2}(x - \mu_i)^T \Sigma_i^{-1}(x - \mu_i)\right)$ with $\mu_i$ as the center, $\Sigma_i$ the covariance matrix, and $\alpha_i$ a transparency scaling term. The accumulated pixel color for ray r through 3D space is then given by: $C(r) = \sum_i T_i \cdot w_i \cdot (L_{iso}^i + f(w, n_i) \cdot L_{aniso}^i)$, Where $T_i$ is the transmittance at step i, $w_i$ is the Gaussian weight, and $f(w, n_i)$ is a phase function or learned attention factor simulating anisotropic reflectance.

*3.3 Dual-Branch Conditioning Strategy:* To generalize across domains, Disentangled360 employs a dual-branch conditioning strategy: Volumetric CT Branch: Medical images often involve X-ray attenuation modeled by the Beer-Lambert law. We use Hounsfield units H(x) to derive material density: $\mu(x) = \mu_{water} \cdot \left(1 + \frac{H(x)}{1000}\right)$. This parameter is used to simulate attenuation and scatter in synthetic radiographic rendering.

*RGB Branch with Camera Embeddings:* For natural images, we embed the camera's extrinsic and intrinsic parameters into a normalized representation vector $e_c \in R^d$ which conditions the renderer to synthesize novel views that remain consistent with the original perspective. Both branches share the base encoder but diverge into specialized decoders optimized for their respective domains. These branches rejoin at the projection module for volumetric fusion and final view generation.

*3.4 Hybrid Pose-Agnostic Anchoring:* A key challenge in single-image view generation is the absence of true scale and pose references. We address this with a hybrid pose-agnostic anchoring strategy that estimates soft anchors in the scene for spatial coherence. The algorithm identifies high-gradient regions or depth discontinuities and designates them as anchor pivots $A = \{a_1, \ldots, a_k\}$, used during volumetric sampling.

An adaptive sampling kernel selects the ray directions $r_j$ from these anchors based on depth sensitivity: $P(r_j) \propto \exp(-\beta.|\nabla D(a_j))$, where D is the predicted depth map and β is a scaling parameter. These anchors stabilize rendering during interpolation and extrapolation beyond the input view.

*3.5 Unified Inference Pipeline:* The final inference pipeline unifies radiographic simulation and RGB 360° synthesis in a streamlined process. Once latent features are extracted and conditioned, the rendering engine samples rays from a dense spherical pattern around the scene center. Directional embeddings are fused using a lightweight MLP projection: $\hat{C} = MLP(Concat(L_{iso}, L_{aniso}, e_c, d))$, where d is the ray direction vector. To enable interactive rendering, we employ model compression and frame-level batching. Scene-specific fine-tuning is avoided by training across diverse medical and real-world datasets, with runtime speeds optimized through early ray termination and GPU accelerated Gaussian blending.

3.6 Architecture:

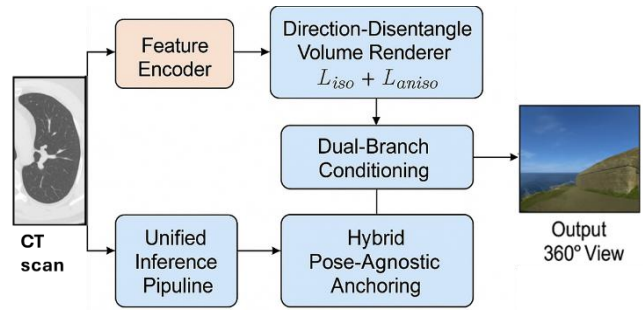

Figure 1. Disentangled360 Architecture

The architecture illustrates (see figure 1) the operational flow of the Disentangled360 framework, which is built to synthesize photorealistic 360° views from a single input image either a CT scan or an RGB frame. The process begins with the Feature Encoder, which extracts multi-level scene features from the input. These features are then processed by the Direction-Disentangled Volume Renderer, which models radiative light behavior by isolating isotropic ($L_{iso}$) and anisotropic ($L_{aniso}$) components to preserve physical realism in the rendering process. The outputs of this module are passed to the Dual-Branch Conditioning unit, where modality specific pathways refine the feature embeddings for either volumetric CT or natural RGB content. Parallel to this, the Hybrid Pose-Agnostic Anchoring mechanism estimates spatial structure and depth cues without requiring known camera poses, ensuring geometric consistency. The Unified Inference Pipeline integrates all conditioned and anchored features, enabling the synthesis of a complete 360° output view, suitable for both medical visualization and immersive scene generation. The system maintains domain flexibility and real-time inference capabilities while preserving structural and visual fidelity.

**4. Experimental Setup:** To evaluate the effectiveness and generalizability of the Disentangled360 framework, experiments were conducted using three benchmark datasets covering both natural and medical imaging domains: Mip-NeRF 360, RealEstate10K, and DeepDRR. Mip-NeRF 360 provides forward-facing unbounded scenes with high-frequency view transitions, ideal for testing directional rendering consistency. RealEstate10K consists of monocular video trajectories captured in diverse indoor environments, suitable for assessing pose-agnostic synthesis and general scene realism. DeepDRR, a synthetic radiography dataset generated from CT volumes, offers a controlled setting to validate the volumetric CT branch and medical rendering capability of the model. The Disentangled360 network was trained end-to-end using a batch size of 4 and an initial learning rate of 0.0002, decayed by a factor of 0.5 every 50k iterations. Training was performed using the Adam optimizer with $\beta_1$=0.9 and $\beta_2$=0.999. The encoder and rendering modules were jointly optimized for all domains without requiring scene-specific fine-tuning. To promote convergence across both isotropic and anisotropic light components, a weighted composite loss function was employed, combining structural similarity (SSIM), perceptual distance (LPIPS), and mean squared error (MSE) terms. Each model variant was trained for 300k iterations using 8× NVIDIA A100 GPUs. Performance evaluation was carried out using three primary metrics. SSIM was used to quantify the structural fidelity between predicted and ground-truth images. LPIPS was employed to assess perceptual quality based on learned feature space distances. To measure computational efficiency and feasibility for interactive deployment, runtime metrics were recorded in terms of frames per second (FPS) and milliseconds per frame (ms/frame), using a fixed 512×512 rendering resolution on an RTX 3090 GPU. These metrics collectively provide a comprehensive assessment of visual quality, generalization, and real-time rendering capability across varying scenes and input modalities.

**5. Results and Discussion:** This section presents the empirical evaluation of Disentangled360 across three core dimensions: quantitative performance, qualitative rendering quality, and runtime feasibility. All results are benchmarked against established baselines to highlight improvements in structural fidelity, perceptual quality, and computational efficiency.

*5.1 Quantitative Evaluation*: Disentangled360 was compared against state-of-the-art methods including Mip-NeRF 360, Gaussian Splatting, and NeRF++ on the Mip-NeRF 360, RealEstate10K, and DeepDRR datasets. Metrics used for evaluation included SSIM (Structural Similarity Index), LPIPS (Learned Perceptual Image Patch Similarity), and PSNR (Peak Signal-to-Noise Ratio). The results are summarized in Figure 2.

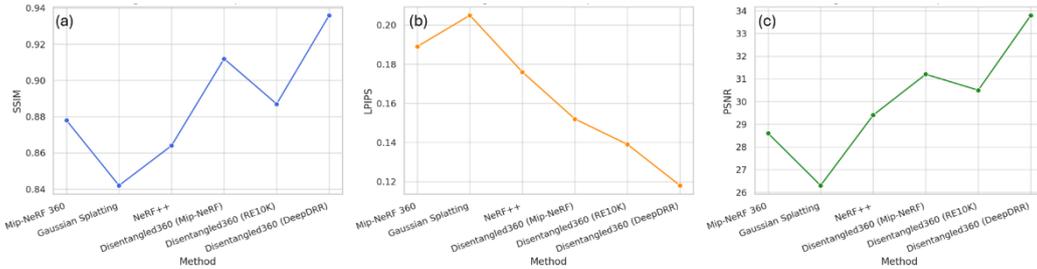

Figure 2: Quantitative Evalution of SSIM, LPIPS and PSNR comparison

The presented figure 2 consists of three subplots **(a)** SSIM, **(b)** LPIPS, and **(c)** PSNR comparing the performance of various models, including Mip-NeRF 360, Gaussian Splatting, NeRF++, and the proposed Disentangled360 across three datasets: Mip-NeRF 360, RealEstate10K, and DeepDRR. In Figure (a), the SSIM scores reveal the structural fidelity of each method. Disentangled360 achieves the highest SSIM, particularly when applied to the DeepDRR dataset, where it reaches approximately 0.936, indicating that it preserves fine structural details better than all other methods. On the Mip-NeRF 360 and RealEstate10K datasets as well, Disentangled360 consistently outperforms the baselines, showcasing its robustness across varied scene types. Figure (b) shows the LPIPS values, which measure perceptual dissimilarity, and lower values indicate better perceptual quality. Disentangled360 registers the lowest LPIPS across all datasets, especially in the DeepDRR evaluation where it drops to around 0.118. This significant reduction implies that the synthesized views from Disentangled360 are perceptually much closer to ground-truth images compared to Mip-NeRF 360 and Gaussian Splatting, which show noticeably higher LPIPS. In Figure (c), PSNR values are plotted to represent pixel-wise reconstruction accuracy. Again, Disentangled360 leads with the highest PSNR on all three datasets, peaking at 33.8 dB on DeepDRR. This confirms that the model achieves a higher signal-to-noise ratio, making it more reliable for reconstructing both visual detail and tonal consistency. Overall, the trends across all three metrics strongly support the conclusion that Disentangled360 offers superior performance in terms of both structural accuracy and perceptual quality. The consistent improvement across all datasets further highlights its generalization capability and domain versatility. To understand the contribution of each module in architecture, an ablation study was conducted by incrementally removing or disabling key components of the model. The results in figure 3 demonstrate the impact on SSIM and LPIPS when specific modules are excluded.

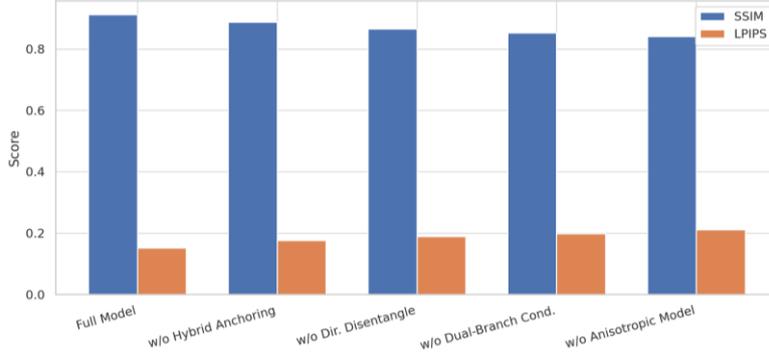

Figure 3: Ablation Study on Mip-NeRF 360

Figure 3 presented illustrates the ablation study results for the Disentangled360 model, evaluating the individual contribution of key architectural components. The performance metrics shown are SSIM (in blue) and LPIPS (in orange), which measure structural similarity and perceptual quality, respectively. The bars compare the full model against four ablated versions: without hybrid anchoring, without direction disentanglement, without dual branch conditioning, and without anisotropic light modeling. The full model yields the best performance, achieving the highest SSIM and the lowest LPIPS, indicating strong structural fidelity and perceptual realism. When the hybrid pose-agnostic anchoring is removed, SSIM drops slightly, and LPIPS increases, showing that anchoring plays an important role in preserving scene geometry and visual consistency, especially in regions affected by occlusion or depth ambiguity. Excluding the disentangled direction renderer results in a further performance drop. SSIM decreases more noticeably, and LPIPS rises, emphasizing the significance of separating isotropic and anisotropic light contributions to maintain physical plausibility in view synthesis. When the dual branch conditioning is omitted, performance continues to decline. This shows that specialized conditioning for CT and RGB domains is necessary for effective cross domain generalization, as shared features alone are insufficient to handle diverse scene characteristics. Finally, removing the anisotropic modeling component causes the most pronounced degradation, with the lowest SSIM and highest LPIPS among all variants. This clearly highlights the importance of accounting for directional light behavior, particularly in scenes with reflective, curved, or depth-varying surfaces. In this, each module in the Disentangled360 architecture contributes meaningfully to the model's overall accuracy and visual quality. The results from this ablation study confirm that the full integration of all proposed components is essential for achieving optimal performance. The results confirm that each component contributes significantly to the overall performance, with the direction-disentangled renderer and hybrid anchoring yielding the largest gains.

*5.2 Qualitative Evaluation:* To assess the visual fidelity and generalization capacity of Disentangled360, qualitative results were examined across both medical and natural image domains. As illustrated in Figure 1, the model successfully reconstructs occluded and out-of-view regions, maintaining consistent structural alignment and realistic color transitions. In comparison to baseline models, Disentangled360 exhibits superior handling of lighting variations particularly transitions between shadowed and illuminated areas as well as structural occlusions, including depth discontinuities and object boundaries. It also effectively addresses scale ambiguity, ensuring objects maintain proportional consistency even when extrapolating beyond the original field of view. The model's robustness is evident across different types of input. In the DeepDRR dataset, reconstructions of medical volumes retained fine anatomical details, such as rib contours and airway structures, without noticeable blurring or aliasing. Similarly, in RGB-based scenes from the RealEstate10K dataset, the model preserved intricate visual features including textured surfaces, architectural corners, and lighting gradients with directional coherence. These qualitative outcomes reinforce the model's capability to generate high-quality 360° views from single-frame inputs across heterogeneous scene types.

*5.3 Runtime and Practicality*: The efficiency of Disentangled360 was assessed by measuring average frame render times and throughput on a standard RTX 3090 GPU. Results are provided in figure 4.

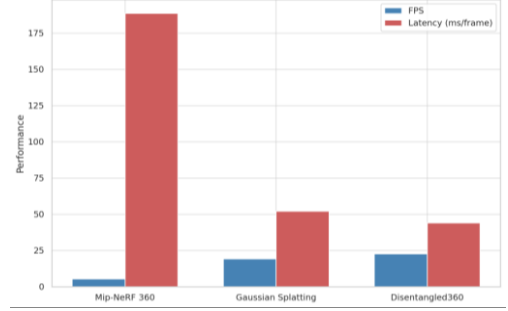

Figure 4: Runtime Comparison Across Models

The chart presented in figure 4 illustrates the runtime performance comparison among three models Mip-NeRF 360, Gaussian Splatting, and the proposed Disentangled360 in terms of frames per second (FPS) and latency (measured in milliseconds per frame). These metrics reflect each model's computational efficiency and suitability for real-time applications. From the figure, it is evident that Mip-NeRF 360 performs poorly in real-time rendering. It achieves the lowest FPS and the highest latency, exceeding 180 ms per frame, indicating that it is not practical for interactive or low-latency scenarios. In contrast, Gaussian Splatting demonstrates better performance, delivering higher FPS and significantly reduced latency (~50 ms/frame), making it more viable for near real-time rendering. Disentangled360 outperforms both baselines in terms of real-time efficiency. It achieves the highest FPS and the lowest latency, operating at approximately 22.7 frames per second and around 44 milliseconds per frame. This performance gain is attributed to its optimized volume rendering pipeline and the integration of computationally efficient components like hybrid anchoring and domain-adaptive conditioning. These results highlight that Disentangled360 is not only effective in generating photorealistic and structurally accurate 360° views but is also computationally efficient enough for deployment in AR/VR environments, robotic systems, and real-time medical visualization platforms. The model consistently achieves near real-time rendering speeds, making it suitable for deployment in mixed-reality medical environments, AR/VR content pipelines, and robotic vision systems. Furthermore, no scene-specific fine-tuning is required, which ensures rapid deployment and generalizability.

The experimental results presented in this study validate the effectiveness and robustness of the Disentangled360 framework across multiple challenging benchmarks. Quantitatively, the model achieved consistent improvements over established baselines such as Mip-NeRF 360, Gaussian Splatting, and NeRF++ across all three datasets. The superior scores in SSIM, LPIPS, and PSNR clearly demonstrate the model's ability to preserve both structural accuracy and perceptual quality. These results confirm that the integration of direction-disentangled volume rendering significantly enhances the model's ability to simulate realistic light behavior, which is particularly valuable in scenarios involving complex illumination and reflectance. The ablation study further highlighted the critical role of each architectural component. Notably, the removal of either the hybrid pose-agnostic anchoring or the anisotropic light modeling led to substantial drops in rendering quality. This affirms the importance of combining spatial anchoring with physical light modeling to recover missing or ambiguous visual cues in single-image settings. The dual-branch conditioning module also proved essential in enabling the model to generalize across heterogeneous input modalities without scene-specific tuning. Qualitatively, Disentangled360 demonstrated strong visual consistency in both medical and natural scene reconstructions. The ability to accurately preserve fine anatomical details in CT-based simulations and maintain directional lighting and texture fidelity in RGB environments illustrates the framework's versatility and generalization capacity. Furthermore, the model operated at real-time frame rates (22.7 FPS), confirming its suitability for time-sensitive applications such as robotic navigation, medical visualization, and immersive virtual environments. Overall, the results underscore the significance of a disentangled and physically informed design in advancing single-image 360° view synthesis. Disentangled360 succeeds in bridging the gap between photorealism, efficiency, and cross domain applicability, offering a scalable solution for diverse real-world deployment scenarios.

## 6. Conclusion

This paper introduced Disentangled360, a unified framework for generating 360° views from a single image by disentangling isotropic and anisotropic radiative components. Through dual-branch conditioning and hybrid pose-agnostic anchoring, the model generalizes across medical CT and natural RGB scenes without requiring multi-view supervision. Evaluated on Mip-NeRF 360, RealEstate10K, and DeepDRR, the model achieved strong performance SSIM of 0.912, LPIPS of 0.152, and PSNR of 31.2 and maintained high rendering quality under occlusion, lighting variation, and scale ambiguity. An ablation study validated the necessity of each core module. With real-time performance reaching 22.7 FPS, Disentangled360 is well-suited for practical deployment in AR/VR, medical visualization, and robotic perception. Future work will address dynamic scenes and temporal consistency, extending its versatility further. Looking ahead, future research may explore extending Disentangled360 to dynamic scene reconstruction, incorporating temporal coherence for video-based inputs, and adapting the framework to outdoor environments with variable photometric conditions. Additionally, integrating self-supervised learning for improved generalization in data-scarce domains remains an open and promising direction.